%% file: main.tex
\documentclass{article}

\usepackage[preprint]{neurips_2022}

\usepackage[utf8]{inputenc} 
\usepackage[T1]{fontenc}    
\usepackage{url}            
\usepackage{booktabs}       
\usepackage{amsfonts}       
\usepackage{nicefrac}       
\usepackage{microtype}      
\usepackage{xcolor}         
\input{math_commands}

\usepackage{hyperref,cleveref}
\usepackage{multirow}
\usepackage{graphicx}
\usepackage{authblk}

\Crefname{equation}{Eq.}{Eqns.}
\Crefname{figure}{Fig.}{Figs.}
\creflabelformat{equation}{(#2#1#3)}

\title{i-Code: An Integrative and Composable Multimodal Learning Framework}

\author{
\parbox{\linewidth}{\centering
\textbf{Ziyi Yang\thanks{Co-first authors.}$\;\:$, Yuwei Fang$^{*}$, Chenguang Zhu, Reid Pryzant, Dongdong Chen}, 

\textbf{Yu Shi, Yichong Xu, Yao Qian, Mei Gao, Yi-Ling Chen, Liyang Lu, }

\textbf{Yujia Xie, Robert Gmyr, Noel Codella, Naoyuki Kanda, Bin Xiao, }

\textbf{Lu Yuan, Takuya Yoshioka, Michael Zeng, Xuedong Huang}

Microsoft Azure Cognitive Services Research

\texttt {\{ziyiyang, yuwfan, chezhu\}@microsoft.com}}
}

\begin{document}

\maketitle

\begin{abstract}
Human intelligence is multimodal; we integrate visual, linguistic, and acoustic signals to maintain a holistic worldview. Most current pretraining methods, however, are limited to one or two modalities. We present i-Code, a self-supervised pretraining framework where users may flexibly combine the modalities of vision, speech, and language into unified and general-purpose vector representations. In this framework, data from each modality are first given to pretrained single-modality encoders. The encoder outputs are then integrated with a multimodal fusion network, which uses novel attention mechanisms and other architectural innovations to effectively combine information from the different modalities. The entire system is pretrained end-to-end with new objectives including masked modality unit modeling and cross-modality contrastive learning. Unlike previous research using only video for pretraining, the i-Code framework can dynamically process single, dual, and triple-modality data during training and inference, flexibly projecting different combinations of modalities into a single representation space. Experimental results demonstrate how i-Code can outperform state-of-the-art techniques on five video understanding tasks and the GLUE NLP benchmark, improving by as much as 11\% and demonstrating the power of integrative multimodal pretraining.

\end{abstract}

\section{Introduction}

True humanlike intelligence incorporates information from a variety of signals and sensory organs \cite{schank1975scripts}. This implies that intelligent systems should be integrative, incorporating signals from all available modalities \cite{lake2017building}. In many practical data regimes this corresponds to the modalities of vision (V), language (L), and speech/audio (S). Although there has been tremendous progress in making models to understand one modality \citep{bert, swin, hubert, wavlm} or two modalities \citep{vl-bert, vilbert, visualbert, clip, align, florence}, it is a non-trivial task to extend these successes to a three-modality system which can simultaneously interpret vision (V), language (L) and speech (S). 

One important difficulty of three-modality pretraining is that it requires enormous amounts of three-modality data like captioned videos, which is often several orders of magnitude smaller than the available single- or dual-modality data. For instance, the largest annotated video dataset at the time of writing consists of 180M clips \citep{merlot}, while the largest available image-caption dataset has 900M pairs \citep{florence}.

To address this problem, we propose two solutions. First, in addition to three-modality videos, we leverage large-scale dual-modality data, e.g., images with captions (V+L), speech with transcripts (S+L) and video narrations (V+S). This greatly expands the size and diversity of data that the model is exposed to while covering all three target modalities. 
Second, instead of building a standalone model from scratch, we propose a fusing architecture that can readily adopt contextual outputs from state-of-the-art single-modality encoders proposed by the community. 
To this end, we propose i-Code, where i stands for integrative multimodal learning. In i-Code, we cultivate an effective fusing module which integrates the outputs of the single-modality encoders and conducts cross-modality understanding to get a final prediction. To design the best fusing architecture, we experiment with variations on the self-attention mechanism inside the transformer architecture, including mechanisms that cross and merge the attention scores of different modalities.

i-Code is then pretrained on double and triple-modality data using a variety of self-supervision objectives, including: i) masked unit modeling, where all input signals are converted into discrete tokens, and the goal is to predict the correct tokens for the masked units of each modality; ii) contrastive learning, where two input modalities are provided and the model predicts whether the given signals come from the same triple (or pair) in the training data. We thoroughly evaluate i-Code on multiple multimodal benchmarks. The experimental results demonstrate the effectiveness of the proposed multimodal pretraining framework. 
Finetuning i-Code beats out state-of-the-art algorithms across six multimodal datasets and the GLUE NLP benchmark, improving over the previous best by as much as 11\%.

\section{Related Work}
Jointly co-learning vision and language representations is an active area of multimodal research. One category of models follows a two-tower architecture with independent encoders for two modalities \citep{clip, align, florence, splat}. In this case, multimodality fusion is achieved via a projection layer which is added to the single-modality encoder. These models are pretrained on image-caption or speech-text data with contrastive learning loss functions. These models exhibit outstanding cross-modality retrieval performance along with zero and few-shot prediction performance. Another body of research seeks to achieve cross-modality interaction via a shared encoder \citep{vit, vilbert, vl-bert, visualbert}. For example, in VL-BERT \citep{vl-bert}, image detection features and language token embeddings are inputted together into a transformer encoder, and the pretraining task is to predict the masked tokens based on the detection features and language contexts. Video-language learning has also been an active research area \citep{vivit, merlot, decembert, howto100m}, where models are trained on videos frames and automatic speech recognition (ASR) transcripts.

Recently, there has been increasing research on modeling the multimodal components of video data: textual transcripts, visual frames, and audio waveform, etc \citep{reserve, vatt, alayrac2020self}. For example, VATT \citep{vatt} builds a transformer encoder on top of projections of raw input data from videos (3D RGB voxels, waveforms and token embeddings) and does not use single-modality encoder. In i-Code, we demonstrate that leveraging state-of-the-art single modality encoders for multimodal learning can effectively boost the multimodal model performance. The i-Code framework also extends the pretraining data from videos to dual-modality datasets.

\section{Large-scale Multimodal Pretraining Data}

To facilitate effective multimodal learning, we collect large-scale multimodal data for pretraining. We collect two types of data: three-modality video and dual-modality datasets.

Video is a large-scale data resource that contains all three modalities and is widely available on public streaming platforms. 
We choose the recently published video dataset YT-Temporal-180M \citep{merlot} because of its great diversity in video corpus topics, high quality filtering and selection, and large-scale quantity. We collect 180 million video clips in the YT-Temporal-180M dataset, using the provided videos IDs and time stamps. On average, each clip is 7.8 seconds. For each clip, we evenly sample 8 video frames as the visual inputs. For speech, the raw waveforms of audios are extracted to be further processed by the downstream speech encoder. Each clip also comes with a textual script that has been carefully denoised from the original ASR transcripts. 
However, misalignment between the frames and transcripts is a concerning and common issue in video data \citep{decembert, howto100m}: narration transcripts can be irrelevant or temporally misaligned with the visual frames. To alleviate this issue, we generate the caption for the high-resolution mid-frame of each clip with the captioning API of Azure Cognitive Services, to augment the video dataset. More details on how we leverage the captions can be found in \Cref{sec:ctrs}.

As high-quality three-modality videos are limited in size, we also resort to dual-modality datasets for pretraining, which have been widely used in applications such as visual-language representation learning \citep{clip, align, florence}, zero-shot cross-modality generation \citep{dall-e}, automatic speech recognition (ASR) and text-to-speech (TTS) \citep{huang2019voice}. i-Code leverages the following dual-modality datasets during pretraining:

\begin{itemize}
  \item \textbf{Visual-Language}. We use 72.8 million image-caption pairs from the pretraining data of the Florence computer vision foundation model \citep{florence}. Data are collected with a programmatic data curation pipeline from the Internet, then selected and post-filtered \citep{florence}.
  
  \item \textbf{Language-Speech}. We use internal 75k-hour transcribed English speech data. This dataset, containing 63.2 million transcript-speech pairs, is diverse in scenarios, including Cortana, far-field speech, and call center.
  
  \item \textbf{Visual-Speech}. For visual and speech pair datasets, we turn to the Spoken Moments in Time (SMiT), a video-narration dataset. SMiT comprises 500k spoken captions each of which depicts a broad range of different events in a short video \citep{smit}. 
\end{itemize}

To the best of our knowledge, this is the first time that paired datasets have been used to train vision-language-speech models. In the experiment section, we compare the performance of models pretrained with paired and video datasets, respectively. We discover that combining both types of datasets can further boost the model's performance.

\section{The i-Code Multimodal Framework}
In this section, we introduce the overall model architecture of i-Code and how we pretrain i-Code on the aforementioned large-scale multimodal datasets in a self-supervised manner.

\subsection{Model Architecture}
i-Code consists of four modules. The first three modules are single-modality encoders (one for vision, language, and speech). The last module is a modality fusion network. The raw input for each modality is fed into its corresponding single-modality encoder, then all encoded inputs are fed through a linear projection layer and integrated with the modality fusion network. Due to this architecture design, i-Code can process various kinds of inputs: single-modality inputs, any combination of two modalities, and all three modalities together.

Instead of training each single-modality encoder from scratch, we design our framework to be modular: any pretrained model can be swapped in to fill the role of a single-modality encoder. This provides the fusion network with high-quality contextual representations for more effective multimodal understanding. We opt to leverage state-of-the-art models for each modality:

\textbf{Language Encoder.} 
We use the recently published DeBERTa V3 base \citep{deberta} as the language encoder. This pretrained language model with 183 million parameters has a disentangled attention mechanism that has helped it achieve record-breaking performance on the GLUE and SuperGLUE NLP benchmarks.

\textbf{Vision Encoder.} We adopt CoSwin Transformer \citet{florence} as the vision encoder. To enable i-Code to process both images and sequence of frames (video), we instantiate a video CoSwin transformer from a pretrained CoSwin Transformer \citet{florence} as the vision encoder, following the procedure in \citet{videoswin}. The video CoSwin transformer has 91 million parameters.

\textbf{Speech Encoder.} Recently, there has been significant advances in learning speech representations through diverse network architectures \citep{wav2vec, wav2vec2, hubert, wavlm}. To leverage these state-of-the-art techniques in speech representation learning, we use a pretrained WavLM-large model \citep{wavlm} of 315 million parameters as the speech encoder. WavLM contains a temporal convolutional encoder to featurize the input speech waveform, followed by a transformer encoder.

It is worth noting that the i-Code framework is integrative and composable such that other single-modality encoders can also be used besides these three mentioned above. For example, we experiment with another speech encoder HuBERT and include the results in \Cref{sec:speech-encoder}.

\textbf{Multimodal Fusion Module.} Features extracted by each single-modality encoder are then projected to the fusion network's hidden dimension by a 1-layer feed-forward network. The projected features are input to the modality fusion network to generate integrative multimodal representations. Since positional information is already incorporated by single-modality encoders, we do not use positional embeddings in the fusion module.

\begin{figure*}[thbp]
\centering
\includegraphics[width=\linewidth]{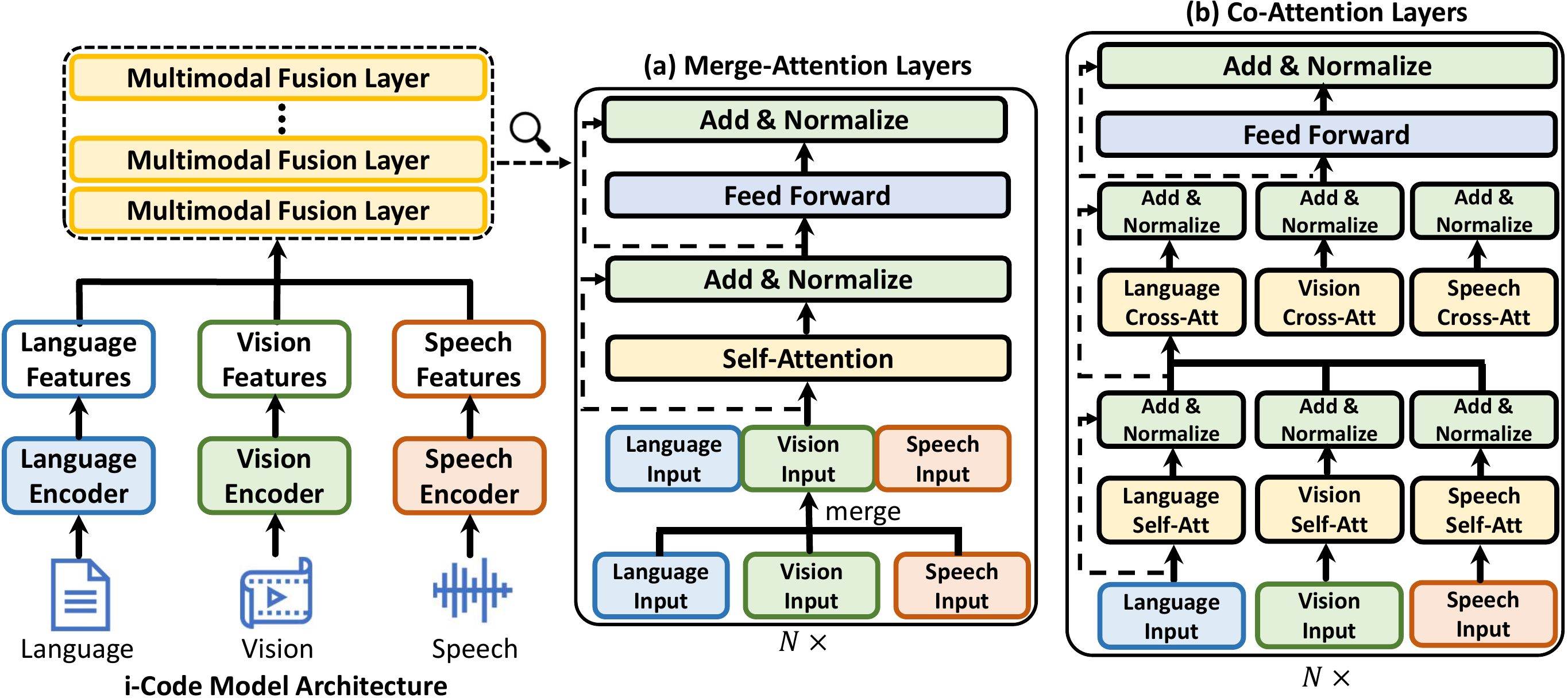}
\caption{Left: The overall model architecture of i-Code. Right: the attention and feed-forward operation in a fusion network layer with (a) merge-attention and (b) co-attention. For simplicity, we only draw the residual connection of the language modality. \label{fig:model}}
\end{figure*}

The backbone of the fusion network is a transformer encoder, where each layer conducts cross-modality attention, forward projection, and layer normalization. 
To facilitate more effective cross-modality understanding, we explore two variations on the traditional attention mechanism: merge-attention and co-attention, as illustrated in \Cref{fig:model}.

\textit{Merge-attention.} In this mode, different modalities share the same attention parameters. To help the fusion module distinguish between different modalities, an identification embedding unique to each modality is added to the projected features (on all temporal and spatial dimensions). Those projected features from different modalities are concatenated together (the temporal and spatial dimensions are flattened for visual inputs) and fed into the fusion network, where each layer is the same as the classical transformer encoder layer \citep{vaswani2017attention}.

\textit{Co-attention.} In this mode, each Transformer layer first conducts self-attention among features of each individual modality, with modality-specific attention parameters. For example, let the language, vision and speech outputs from a previous transformer layer be $X_L$, $X_V$ and $X_S$. Now, we can write a single attention head focusing on the language modality as: $$X_{L}^{\text{self}} = \text{Self-Attention-Language}(Q_L, K_L, V_L),$$
where query $Q_L=X_L W_L^Q$, key $K_L=X_L W_L^K$, value $V_L=X_L W_L^V$; $W_L^Q$, $W_L^K$, and $W_L^V$ are modality-specific attention matrices. The self-attention sublayer is followed by a cross-modality attention:
$$X_{L}^{\text{cross}} = \text{Cross-Attention-Language}(Q_L^{\text{cross}}, K_L^{\text{cross}}, V_L^{\text{cross}}),$$

where $Q_L^{\text{cross}}=X_L^{\text{self}} W_{Lc}^Q$,
$K_L^{\text{cross}}=[X_V^{\text{self}}, X_S^{\text{self}}] W_{Lc}^K$,
$V_L^{\text{cross}}=[X_V^{\text{self}}, X_S^{\text{self}}] W_{Lc}^V$; $W_{Lc}^Q$, $W_{Lc}^K$, and $W_{Lc}^V$ are cross-attention parameters of the language modality. For a fusion network module with merge-attention, we use 6 transformer encoder layers with hidden size $768$ and the fusion module has 154 million parameters. For the co-attention fusion module, to keep its model size close to the merge-attention one, we use 3 layers and the same hidden size, ending up with 163 million parameters. The parameters in the fusion module are randomly initialized in pretraining and are not instantiated from pretrained checkpoints.

\subsection{Pretraining i-Code}
In this subsection, we introduce how we pretrain i-Code. We first discuss the multimodal pretraining objectives: masked units modeling and cross-modality contrastive learning. Then we introduce the optimization and training details.

\subsubsection{Masked Units Modeling}

The first group of self-supervised pretraining objectives are the masked unit modeling objectives for each modality.

\textbf{Masked Language Modeling (MLM).} Masked Language Modeling (MLM) has achieved remarkable success in self-supervised learning for both language \citep{bert} and vision-language pretraining \citep{meter}. During pretraining, we mask out 30\% of the text tokens\footnote{Similarly to BERT pretraining, 10\% of the masked tokens are replaced with random token, and 10\% of the time we keep the tokens unchanged and 80\% are replaced with the MASK token.}. The task is to predict the masked tokens and the loss $\gL_{\mathrm{MLM}}$ is the cross entropy between the ground-truth and the predicted token indices.

\textbf{Masked Vision Modeling (MVM).} For vision self-supervised learning, we adopt the consistent high-level strategy as masked language modeling. We convert vision inputs to discrete tokens, mask out regions in the inputs images, and maximize the cross-entropy between the prediction and ground-truth tokens of the masked language out regions. Given a sequence of frames, we leverage PeCo \citep{peco}, a state-of-the-art visual vector quantized variational Autoencoder (VQ-VAE), to discretize each frame to tokens. For masking, we adopt the 3D tube-masking strategy proposed in \citet{bevt} to mask image regions across the temporal dimension, where the masking patch ratio for one frame is 50\%. We introduce the details in \Cref{apd_sec:mvm}.

\textbf{Masked Span Modeling (MSM).} We discretize the speech utterance into a sequence of tokens with a pretrained speech quantizer model, where we use the quantizer of wav2vec 2.0 \citep{wav2vec2}. We use the same masking strategy as in HuBERT \citep{hubert} and wav2vec 2.0 \citep{wav2vec2}, where $p\%$ of the time steps are randomly selected as start indices, and the next $l$-step span is masked \citep{hubert}. We follow the default setting of pretraining WavLM and HuBERT by choosing $l=10$ and $p=8$. The MSM loss $\gL_{\mathrm{MSM}}$ is the cross-entropy between the predicted and ground-truth labels.

\subsubsection{Cross-Modality Contrastive Learning}
\label{sec:ctrs}
The second group of pretraining objectives are the cross-modality contrastive learning objectives.
Each single modality input is first encoded by the corresponding encoder and then fed into the multimodal encoder \textbf{individually}. Next, we mean-pool each group of single-modality embeddings. For language and speech, the multimodal encoder outputs are averaged along the sequential/temporal dimension. For vision inputs, they are averaged along both temporal and spatial dimensions. We denote the obtained representations for vision, language and speech as ${\vr_v, \vr_l, \vr_s}$ respectively. We then normalize the representations to the unit vector, e.g., $\vu_v = \frac{\vr_v}{\|\vr_v\|_2}$. 

The vision-language contrastive loss $\gL_{vl}$ for a minibatch $\gB$ is then defined as:
\begin{equation}
    \gL_{vl} = \gL_{v2l} + \gL_{l2v},
\label{eq:ctrs}
\end{equation}

where the vision-to-language and language-to-vision contrastive learning objectives are given as:
\begin{equation}
    \gL_{v2l} = -\frac{1}{|\gB|} \sum_{i=1}^{|\gB|} \frac{\exp(\tau_{vl} \innerproduct{\vu_v^{(i)}}{\vu_l^{(i)}})}{\sum_{j=1}^{|\gB|}\exp(\tau_{vl}\innerproduct{\vu_v^{(i)}}{\vu_l^{(j)}})},
    \ 
    \gL_{l2v} = -\frac{1}{|\gB|} \sum_{i=1}^{|\gB|} \frac{\exp(\tau_{vl} \innerproduct{\vu_l^{(i)}}{\vu_v^{(i)}})}{\sum_{j=1}^{|\gB|}\exp(\tau_{vl}\innerproduct{\vu_l^{(i)}}{\vu_v^{(j)}})}.
\label{eq:v2l}
\end{equation}
$\tau_{vl}$ is a learnable parameters to scale the logits of the vision-language pairs and $ \langle  \; , \; \rangle $ denotes the inner product. Similarly, we define the contrastive learning objectives $\gL_{vs}$ and $\gL_{ls}$ for vision-speech and language-speech respectively. To further increase the effective batch size of contrastive learning, we concatenate the batches across all GPUs and employ the recently proposed gradient-cache technique \citep{gradcache}.

For pretraining on videos, captions and ASR transcripts are concatenated as the language input to the visual-language contrastive learning and the language masked language modeling. The final pretraining objective is the weighted sum of the masked units modeling and contrastive learning objectives:

\begin{equation}
\gL = \alpha \gL_{\mathrm{MLM}} + \beta \gL_{\mathrm{MVM}} + \gamma \gL_{\mathrm{MSM}} + \lambda (\gL_{vl} + \gL_{vs} + \gL_{ls}).    
\label{eq:pretrain_loss}
\end{equation}

We experiment with both fixed and learnable combination weights. We do not observe significant differences in downstream performance and empirically find that $\alpha=0.5, \beta=0.6, \gamma=1, \lambda=1$ works well. On either paired datasets or videos, we pretrain for 3 epochs on 72 A100 GPUs, with effective batch size 1728. The learning rate is $2 \times 10^{-5}$ for the fusion module and $10^{-5}$ for modality encoders with 20000 warm-up steps, and the optimizer is AdamW.

\section{Experiments}
In this section, we evaluate i-Code and compare it with previous work on a variety of downstream tasks, including multimodal sentiment \& emotion analysis, multimodal inference, video QA and single-modality tasks. Due to the space limit, we refer readers to the appendix for details on experiment settings, hyper-parameters, and performance standard deviation for each downstream task.

\subsection{Multimodal Sentiment \& Emotion Analysis}
We test i-Code on the largest dataset of multimodal sentiment analysis and emotion recognition to date, CMU Multimodal Opinion Sentiment and Emotion Intensity (CMU-MOSEI) \citep{mosei} that consists of 23,453 utterance videos. It contains two tasks: sentiment analysis and emotion recognition. For sentiment analysis, given a video, the models need to predict the sentiment levels from ``highly negative (-3)'' to ``highly positive (3)'' \citep{mosei}. Evaluation metrics are mean average errors (MAE), correlation (Corr) between the predicted and ground-truth sentiment labels and F1 score. The dataset can also be evaluated as a binary classification task, by grouping sentiment -3 to -1 \footnote{Or to 0, in which case results are presented on the right of ``/'' in column ``Acc-2'' and ``F1'' of \Cref{tab:mosei-senti}.} to one class and 1 to 3 to another.
 
 We test several configurations of i-Code, including fusion attention mechanism (merge- v.s. co-attention) and pretraining data (dual v.s. video dataset, or combined). Results are shown in \Cref{tab:mosei-senti}. i-Code sets the state-of-the-art on this task, e.g., improving 3\% on the correlation. i-Code models trained on the dual dataset exhibit better performance than the one trained on video dataset, and the merge-attention outperforms the co-attention on this dataset. We also explore directly finetuning on the downstream task, without pretraining the fusion module (denoted as ``No Pretrain''). Leveraging the state-of-the-art encoders already shows competitive performance against previous models, even without multimodal pretraining.

\begin{table*}[htbp]
\centering
\caption{Results on CMU MOSEI Sentiment Analysis.}
\begin{tabular}{l|c|c|cccc}
\toprule
\multicolumn{3}{l|}{Model} & MAE $(\downarrow)$ & Corr. & Acc-2 & F1 \\ \midrule
\multicolumn{3}{l|}{MulT \citep{mult}} & 0.591 & 69.4 & -/81.6 & -/81.6 \\
\multicolumn{3}{l|}{ICCN \citep{iccn}} & 0.565 & 71.3 & -/84.2 & -/84.2 \\
\multicolumn{3}{l|}{MISA \citep{misa}} & 0.555 & 75.6 & 83.6/85.5 & 83.8/85.3 \\
\multicolumn{3}{l|}{ScaleVLAD \citep{scalevlad}} & 0.527 & 78.1 & 84.5/86.4 & 84.7/86.3 \\
\multicolumn{3}{l|}{Self-MM \citep{selfmm}} & 0.530 & 76.5 & 82.8/85.2 & 82.5/85.3 \\ \midrule
\multirow{5}{*}{\textbf{i-Code}} & Pretrain Data & Fusion Attention & & & &  \\ \cmidrule{2-7}
 & No Pretrain & Merge & 0.529 & 79.6 & 84.7/86.8 & 84.8/86.5 \\
 & Dual & Merge & \textbf{0.502} & \textbf{81.1} & 85.3/\textbf{87.5} & \textbf{85.6/87.4} \\
 & Dual & Co & 0.525 & 80.9 & 84.1/87.1 & 84.3/87.0 \\
 & Video & Merge & 0.519 & 80.8 & \textbf{85.4}/87.3 & 85.5/87.1 \\
 & Dual+Video & Merge & 0.507 & 80.7 & 83.8/87.3 & 84.1/87.2 \\
\bottomrule
\end{tabular}
\label{tab:mosei-senti}
\end{table*}

For emotion recognition, videos are categorized into Ekman emotions \citep{ekman} of \{happiness, sadness, anger, fear, disgust, surprise\}. The evaluation metrics are accuracy, precision, recall and Micro-F1. We evaluate on the unaligned version of the dataset, since the alignment information is not always available in real-world scenarios. Results are listed in \Cref{tab:mosei-emo} and i-Code improves upon previous best models by 4.1\% on accuracy and 3.3\% on F1. The co-attention surpasses the merge-attention. Leveraging dual and video data together for pretraining improves upon using either of both. In the appendix \Cref{sec:alignment}, we show that i-Code even surpasses previous models which had additional access to the alignment information.

\begin{table*}[htbp]
\centering
\caption{Results on CMU MOSEI Emotion Recognition.}
\begin{tabular}{l|c|c|cccc}
\toprule
\multicolumn{3}{l|}{Model} & Acc. & F1 & Precision & Recall \\ \midrule
\multicolumn{3}{l|}{DFG \citep{mosei}} & 38.6 & 49.4 & 53.4 & 45.6  \\
\multicolumn{3}{l|}{MISA \citep{misa}} & 39.8 & 45.0 & 37.1 & 57.1 \\
\multicolumn{3}{l|}{RAVEN \citep{raven}} & 40.3 & 51.1 & 63.3 & 42.9 \\
\multicolumn{3}{l|}{MuIT \citep{mult}} & 42.3 & 52.3 & 63.6 & 44.5 \\
\multicolumn{3}{l|}{HHMPN \citep{hhmpn}} & 43.4 & 52.8 & 59.1 & 47.6 \\
\multicolumn{3}{l|}{TAILOR \citep{tailor}} & 46.0 & 52.9 & \textbf{63.9} & 45.2 \\
\multicolumn{3}{l|}{SIMM \citep{simm}} & 41.8 & 48.4 & 48.2 & 48.6 \\
\multicolumn{3}{l|}{ML-GCN \citep{mlgcn}} & 43.7 & 52.4 & 57.3 & 48.2 \\ \midrule
\multirow{5}{*}{\textbf{i-Code}} & Pretrain Data & Fusion Attention & & & & \\ \cmidrule{2-7}
 & No Pretrain & Merge & 49.2	& 54.6 & 50.3 & 59.8 \\
 & Dual & Merge & 49.4  & 55.4 & 49.4 & \textbf{63.0} \\
 & Dual & Co & \textbf{50.2} & \textbf{56.2} & 50.7 & \textbf{63.0} \\
 & Video & Merge & 49.4 & 55.3 & 49.6 & 62.4 \\
 & Dual+Video & Merge & 49.8 & 56.0 & 50.8 & 62.1 \\
\bottomrule
\end{tabular}
\label{tab:mosei-emo}
\end{table*}

We then test on a humor detection dataset, UR-FUNNY \citep{urfunny}. In this binary classification dataset, given a video clip with subscripts, video frames and sound, the task to predict whether this clip will lead to immediate laughter. Baseline models include those that also leverage three-modality inputs, e.g., Bi-Bimodal-Fusion Network \citep{bbfn}, Low-rank Matrix Fusion (LMF, \citet{lmf}), MultiBench \citep{multibench} and Tensor Fusion Network (TFN, \citet{tfn}). Due to space limitations, the i-Code model pretrained with ``dual datasets'' is abbreviated as ``D'', ``videos'' as ``V'', ``no pretraining'' as ``NP'', the merge-attention fusion network as ``M'', and the co-attention fusion network as ``C''. E.g., the i-Code model trained on dual datasets with the co-attention is denoted as ``\textbf{i-Code D+C}''. i-Code outperforms the previous best model by 7.5\% and video pretraining shows the best performance.

\begin{table*}[htbp]
\centering
\caption{Results on the UR-FUNNY dataset.}
\resizebox{\linewidth}{!}{
\begin{tabular}{l|ccccccccccc}
\toprule
\multirow{2}{*}{Model} & \multirow{2}{*}{\shortstack{\textbf{i-Code}\\D+M}} &
\multirow{2}{*}{\shortstack{\textbf{i-Code}\\NP+M}} & 
\multirow{2}{*}{\shortstack{\textbf{i-Code}\\D+C}} & 
\multirow{2}{*}{\shortstack{\textbf{i-Code}\\V+M}} &
\multirow{2}{*}{MulT}  &  \multirow{2}{*}{MISA}  &  \multirow{2}{*}{MultiBench} & \multirow{2}{*}{BBFN}  &  \multirow{2}{*}{LMF} & \multirow{2}{*}{TFN}  \\
 &  &  &  &  &  &  &  &  & \\
\midrule
Acc. & 76.94 & 73.16 & 77.00 & \textbf{79.17} & 70.55 & 70.61 & 66.7       & 71.68 & 67.53 & 68.57 \\
\bottomrule
\end{tabular}}
\label{tab:urfunny}
\end{table*}

\subsection{Multimodal Inference}
 To better assess how i-Code reasons across modalities, we evaluate i-Code on a multimodal inference dataset VIOLIN \citep{violin}. In VIOLIN, the input is a video clip from a television show. This clip consists of frames $V$, aligned subtitles $T$ and sound $S$. The clip is paired with a text hypothesis $H$ and the task is to decide whether the hypothesis contradicts or entails the video clip. In i-Code evaluations, we append the video subtitles to the text hypothesis $H$, separated by the [SEP] token. The multimodal representation is computed as the average of the outputs from the fusion network. A binary classifier is trained on the ensuing multimodal representation. Results are summarized in \Cref{tab:violin}. Baselines include HERO, Craig.Starr, GVE (a model specialized for video-language entailment, \citet{gve}), and DF-BERT (the best baseline in the original VIOLIN paper that uses detection features from a Faster R-CNN and BERT). i-Code improves upon the previous best model by 3.5\%.

\begin{table}[htbp]
\centering
\caption{Results on the VIOLIN dataset.}
\resizebox{\linewidth}{!}{
\begin{tabular}{l|cccccccccc}
\toprule
\multirow{2}{*}{Model} & \multirow{2}{*}{\shortstack{\textbf{i-Code}\\D+M}} & 
\multirow{2}{*}{\shortstack{\textbf{i-Code}\\NP+M}} &
\multirow{2}{*}{\shortstack{\textbf{i-Code}\\D+C}} &
\multirow{2}{*}{\shortstack{\textbf{i-Code}\\V+M}} &
\multirow{2}{*}{HERO}  & \multirow{2}{*}{Craig.Starr} & \multirow{2}{*}{GVE (C3D)} & \multirow{2}{*}{GVE (ResNet)} & \multirow{2}{*}{DF-BERT} \\ 
 & & & & & & & & \\ \midrule
Acc. & \textbf{72.90} & 71.60 & 72.09 & 72.61 & 68.59 & 69.43       & 68.15     & 68.39 & 67.84 \\
\bottomrule
\end{tabular}%
}
\label{tab:violin}
\end{table}

\subsection{Video Question \& Answering}
Another category of downstream we test on is video question answering (VQA). For VQA the AI system is given a video, which contains frames $\vv$, subtitles $\vt$ and audio $\vs$ (if available), as well as a question $q$. The system needs to choose the right answer from several candidates $\{a_i\}$. Following previous works on VQA \citep{hero, knowit}, we concatenate the question, a candidate answer, and subtitles together (separated by the [SEP] token) as the text input. Then the text input, visual frames and speech waveforms are input together to the i-Code model to average the outputs across modalities as the multimodal representation $\vr_i$ of the pair $\{q, a_i\}$. A projection layer transforms the representation to a logit, and the softmax function is applied on logits from all answer candidates.

The first VQA benchmark is How2QA \citep{hero}, which contains 31.7k video clips collected from HowTo100M \citep{howto100m}. The baseline models on How2QA inlcude HERO \citep{hero}, the 2021 ICCV VALUE winner Craig.Starr \citep{cstar}, DUKG \citep{dukg}, CLIP \citep{clip}, CLIP+SlowFast features and ResNet+SlowFast features \citep{value}. Results on the public test split are listed in \Cref{tab:how2qa}

\begin{table*}[htbp]
\centering
\caption{Results on the HOW2QA dataset.}
\resizebox{\linewidth}{!}{
\begin{tabular}{l|cccccccccc}
\toprule
\multirow{2}{*}{Model} &
\multirow{2}{*}{\shortstack{\textbf{i-Code}\\D+M}} &
\multirow{2}{*}{\shortstack{\textbf{i-Code}\\NP+M}} &
\multirow{2}{*}{\shortstack{\textbf{i-Code}\\D+C}} &
\multirow{2}{*}{\shortstack{\textbf{i-Code}\\V+M}} &
\multirow{2}{*}{Craig.Starr} & \multirow{2}{*}{HERO}    & \multirow{2}{*}{DUKG}  & \multirow{2}{*}{CLIP}  & \multirow{2}{*}{CLIP-SF} & \multirow{2}{*}{ResNet-SF} \\ 
 & & & & & & & & & & \\\midrule
Acc. & 75.41 & 74.41 & 75.52 & \textbf{75.73} & 74.74 & 74.32   & 73.92 & 69.34 & 72.87   & 74.32  \\
\bottomrule
\end{tabular}}
\label{tab:how2qa}
\end{table*}

KnowIT is a knowledge-based VQA dataset with 24,282 human-annotated question-answer pairs \citep{knowit} and each question has 4 candidate answers (audios are unavailable in this dataset). We compare i-Code with DiagSumQA \citep{diagsumqa}, variants of knowledge based VQA models ROCK \citep{knowit} and ROLL \citep{roll}. Results are summarized in \Cref{tab:knowit} and i-Code achieves the state-of-the-art results on the KnowIT dataset.

\begin{table*}[htbp]
\centering
\caption{Results on the KnowIT video Q\&A dataset.}
\resizebox{\linewidth}{!}{
\begin{tabular}{l|ccccccccccc}
\toprule
\multirow{2}{*}{Model} &
\multirow{2}{*}{\shortstack{\textbf{i-Code}\\D+M}} &
\multirow{2}{*}{\shortstack{\textbf{i-Code}\\NP+M}} &
\multirow{2}{*}{\shortstack{\textbf{i-Code}\\D+C}} &
\multirow{2}{*}{\shortstack{\textbf{i-Code}\\V+M}} &
\multirow{2}{*}{DiagSumQA}  & \multirow{2}{*}{ROLL} & \multirow{2}{*}{\shortstack{ROCK \\ concepts}} & \multirow{2}{*}{\shortstack{ROCK \\ image}} & \multirow{2}{*}{\shortstack{ROCK \\ facial}} & \multirow{2}{*}{\shortstack{ROCK \\ caption}} \\
&  &  &  &  &  &  &  &  &  & \\ \midrule
Acc. & \textbf{80.5} & 78.1 & \textbf{80.5} & 80.0 & 78.1 & 71.5 & 65.4 & 65.4 & 65.4 & 63.5 \\
\bottomrule
\end{tabular}}
\label{tab:knowit}
\end{table*}

\subsection{Single Modality}

We further investigate the performance of i-Code on single-modality tasks, e.g., language-only NLP tasks. For example, in \Cref{tab:glue}, we compare i-Code (D+M) against previously published multimodal models as well as several language-specific models with a similar number of parameters (e.g., BERT, RoBERTa, and DeBERTa V3) on GLUE \citep{glue}. Our model i-Code has set a new state-of-the-art for multimodal models by a significant margin of 11\%. Compared to previous language-only models, i-Code also shows stronger performance and outperforms DeBERTa V3 (which the i-Code language encoder is initialized from) on 7 out of 8 tasks as well as overall performance. As indicated by results in \Cref{tab:glue}, language encoders in previous multimodal work, especially V+L models, usually exhibit inferior performance compared to language models on language-only tasks. The performance gap is typically attributed to the inferior quality of language data in multimodal datasets (e.g., image-caption pairs). We speculate that the masked language modeling objective, as well as the language-speech contrastive learning help i-Code close the gap.

\begin{table*}[htbp]
\centering
\caption{Comparison to previous models on language tasks (GLUE). Results are collected from the original papers or \citet{flava}. Best results of multimodal models are in bold and the highest performance of both multi- and single-modal models are underlined.}
\begin{tabular}{l|cccccccccc}
\toprule
Model      & CoLA & SST-2 & RTE  & MRPC      & QQP       & MNLI & QNLI & STS-B & Avg. \\ \midrule
\multicolumn{10}{c}{\textit{Single-modality Language Models}} \\ \midrule
BERT Base       & 52.1 & 93.5  & 66.4 & 88.9      & 71.2      & 84.6 & 90.5 & 85.8 & 79.1 \\
RoBERTa Base   & 63.6 & 94.8  & 78.7 & 90.2 & 91.9 & 87.6 & 92.8 & 91.2 & 86.4 \\
DeBERTa V3 Base & 69.2 & 96.2  & 84.8 & 90.2      & 92.5      & \underline{90.6} & 94.1 & 91.4 & 88.6 \\ \midrule
\multicolumn{10}{c}{\textit{Multimodal Models}} \\ \midrule
UNITER     & 37.4 & 89.7  & 55.6 & 69.3 & 89.2 & 80.9 & 86.0 & 75.3 & 72.9 \\
FLAVA      & 50.7 & 90.9  & 57.8 & 81.4 & 90.4 & 80.3 & 87.3 & 85.7 & 78.0 \\
VisualBERT & 38.6 & 89.4  & 56.6 & 71.9 & 89.4 & 81.6 & 87.0 & 81.8 &  74.5 \\
VL-BERT    & 38.7 & 89.8  & 55.7 & 70.6 & 89.0 & 81.2 & 86.3 & 82.9 &  74.2 \\ 
ViLBERT    & 36.1 & 90.4  & 53.7 & 69.0 & 88.6 & 79.9 & 83.8 & 77.9 &  72.4 \\
CLIP       & 25.4 & 88.2  & 55.3 & 69.9 & 65.3 & 33.5 & 50.5 & 16.0 &  50.5 \\
\textbf{i-Code}      & \textbf{\underline{70.1}} & \textbf{\underline{96.3}}  & \textbf{\underline{85.6}} & \textbf{\underline{91.0}}    & \textbf{\underline{92.6}}      & \textbf{90.5} & \textbf{\underline{94.3}} & \textbf{\underline{91.9}} &  \textbf{\underline{89.0}} \\
\bottomrule
\end{tabular}
\label{tab:glue}
\end{table*}

\section{Analysis}
\textbf{Modality Ablation Study.} We investigate how effective the single modality or combinations of modalities is/are in the multimodal downstream tasks. Take MOSEI Emotion Reconition as example (\Cref{tab:modality_abl}), we find speech to be the most effective single modality, which makes sense considering the emotional quality of human speech \cite{peerzade2018review}. Leveraging dual modalities improves upon using the single-modality, and language-speech is the best-performing of the dual combinations. 
\begin{table*}[htbp]
\centering
\caption{Modality effectiveness on CMU MOSEI Emotion Recognition.}
\begin{tabular}{c|c|c|cccc}
\toprule
Vision & Language & Speech & Acc. & F1 & Precision & Recall \\ \cmidrule{1-7}
\checkmark &  &  & 45.0 & 50.0 & 45.9	& 54.9  \\
 & \checkmark &  & 46.3 & 52.6 & 44.5 & \underline{64.3} \\
 &  & \checkmark & \underline{47.3} & \underline{52.7} & \underline{46.4} & 60.1 \\ \midrule
 \checkmark & \checkmark &  & 49.0 & 54.8	& \underline{49.2} & 61.8 \\
 \checkmark &  & \checkmark & 48.0 & 53.3 & 47.5 & 60.7 \\
  & \checkmark & \checkmark & \underline{49.2} & \textbf{56.1} & 48.8 & \textbf{65.8} \\ \midrule
\checkmark & \checkmark & \checkmark & \textbf{49.4}  & 55.4 & \textbf{49.4} & 63.0 \\
\bottomrule
\end{tabular}
\label{tab:modality_abl}
\end{table*}

\textbf{Effects of V+L+S Pretraining.} As indicated by results above (\Crefrange{tab:mosei-senti}{tab:knowit}), self-supervised pretraining on large-scale unlabeled multimodal datasets (i-Code D+M, i-Code V+M) improves upon without pretraining (i-Code NP+M).

\section{Conclusion}
In this paper we introduced i-Code, a multimodal pretraining framework that jointly learns representations for vision, language and speech. i-Code is integrative and flexible, able to dynamically process one, two, or three modalities at a time for the construction of shared representation spaces. The model leverages novel attention mechanisms and loss functions to effectively combine information from these disparate modalities. We show that pretraining on dual-modality datasets can also yield competitive or even better performance than pretraining on videos, the data resource that previous three-modality models were restricted to. We thoroughly evaluate i-Code across a range of tasks and domains, setting a new state-of-the-art on 5 video understanding tasks and the GLUE NLP benchmark.

\section*{Limitations \& Societal Impact}
Multimodal models can assist people with sensory disabilities, e.g., vision and hearing loss, in understanding and interacting with multimodal media. These models can also be leveraged for detecting hateful and racial speech and meme, where the contents are often multimodal. The same developed technology, however, can be used in video security surveillance. The model's output could exhibit cultural and societal bias inherited from the pretraining data.

\begin{ack}
We would like to thank Lindsey Li, Lijuan Wang, Junjian Xia, Min Gao, Houdong Hu for the help in data collection. We would like to thank our team members from Microsoft Azure Cognitive Services Research for the insightful discussions and comments.
\end{ack}

\bibliography{ref}
\bibliographystyle{ref}

\clearpage
\appendix

\section{Appendix}

\subsection{Masked Vision Modeling (MVM).} 
\label{apd_sec:mvm}
In this subsection, we provide more details for MVM. Denote the visual input as $\vv \in \R^{W\times H \times T \times 3}$, where $W$, $H$ and $T$ represents the the width, height and temporal dimension respectively. We follow \citet{videoswin} to first convert $\vv$ into $\frac{W}{4} \times \frac{H}{4} \times \frac{T}{2}$ patch features, and each patch feature vector is of size $C$. Then we conduct the tube masking strategy proposed in \citet{bevt} on the patch features as follows.

First, we sample the mask positions $\{i \in [0, \frac{W}{4}), j\in [0, \frac{H}{4})\}$ to create a 2D mask (with $50\%$ masking ratio), the number of masked frames $l \in [\frac{T}{2}, T]$ and the starting frame index $t$. Patches at masking positions $\{i, j\}$ in each frame in $[t, t+l]$ are masked out (i.e., the shape of the 3D mask is a ``tube''), by replacing the patch feature with a learnable mask embedding. The masked image patches are then input to the vision encoder to obtain visual features $f_v \in \R^{\frac{W}{32} \times \frac{H}{32} \times \frac{T}{2} \times H_v}$ ($H_v$ is the output dimension of the visual encoder). The visual features are input to the multimodal fusion network, together with features of other modalities, to obtain the the fused modality features $f_v^m \in \R^{\frac{W}{32} \times \frac{H}{32} \times \frac{T}{2} \times H}$, where $H$ denotes the hidden dimension of the fusion module. A MVM head upsamples $f_v^m$ in both spatial and temporal dimensions, through a 3D deconvolutional network $\mathrm{DeConv}$ \citep{bevt}:
\begin{equation}
    f^{\mathrm{MVM}} = \mathrm{DeConv}(f_v^m) \in \R^{\frac{W}{16} \times \frac{H}{16} \times T \times H},
\label{eq:upsample}
\end{equation}

Next, to convert continuous pixel values to discrete tokens as the prediction labels, we leverage PeCo \citep{peco}, a state-of-the-art visual vector quantized variational Autoencoder (VQ-VAE). PeCo first converts an image patch to a latent vector, and the indice of the closest latent code in the codebook is retrieved as the discrete token for that patch.

The predicted probability distribution of the masked position $\vp_{i, j, t} \in \R^{V_c}$ is then given as:
\begin{equation}
    \vp_{i, j, t} = \mathrm{softmax}(W f_{i, j, t}^{\mathrm{MVM}} + b),
\label{eq:mvm_prob}
\end{equation}
where $V_c$ denotes the codebook size of the PeCo VQ-VAE.

The MVM objective is to maximize the probabilities of the ground-truth token $z_{i, j, t}$:
\begin{equation}
    \gL_{\mathrm{MVM}} = -\frac{1}{|N_m|} \sum_{\{i, j, t\}} \log p_{i, j, t}^{z_{i, j, t}},
\label{eq:mvm}
\end{equation}
where $N_m$ is the number of masked visual patches.

\subsection{Results Standard Deviations}

We list the standard deviations of results for i-Code in on CMU-MOSEI dataset \Cref{tab:mosei-senti-apd,tab:mosei-emo-apd}. They are computed from 5 runs with different seeds for parameters initialization in downstream task finetuning.

\begin{table*}[htbp]
\centering
\caption{Results on CMU MOSEI Sentiment Analysis.}
\resizebox{\linewidth}{!}{
\begin{tabular}{l|c|c|cccc}
\toprule
\multicolumn{3}{l|}{Model} & MAE $(\downarrow)$ & Corr. & Acc-2 & F1 \\ \midrule
\multicolumn{3}{l|}{MulT \citep{mult}} & 0.591 & 69.4 & -/81.6 & -/81.6 \\
\multicolumn{3}{l|}{ICCN \citep{iccn}} & 0.565 & 71.3 & -/84.2 & -/84.2 \\
\multicolumn{3}{l|}{MISA \citep{misa}} & 0.555 & 75.6 & 83.6/85.5 & 83.8/85.3 \\
\multicolumn{3}{l|}{ScaleVLAD \citep{scalevlad}} & 0.527 & 78.1 & 84.5/86.4 & 84.7/86.3 \\
\multicolumn{3}{l|}{Self-MM \citep{selfmm}} & 0.530 & 76.5 & 82.8/85.2 & 82.5/85.3 \\ \midrule
\multirow{5}{*}{\textbf{i-Code}} & Pretrain Data & Fusion Attention & & & &  \\ \cmidrule{2-7}
 & No Pretrain & Merge & 0.529$\pm$0.12 & 79.6$\pm$0.3 & 84.7$\pm$0.3/86.8$\pm$0.3 & 84.8$\pm$0.4/86.5$\pm$0.3 \\
 & Dual & Merge & \textbf{0.502}$\pm0.08$ & \textbf{81.1$\pm$0.2} & 85.3$\pm$0.4/\textbf{87.5$\pm$0.3} & \textbf{85.6$\pm$0.3/87.4$\pm$0.2} \\
 & Dual & Co & 0.525$\pm$0.10 & 80.9$\pm$0.3 & 84.1$\pm$0.2/87.1$\pm$0.3 & 84.3$\pm$0.2/87.0$\pm$0.4 \\
 & Video & Merge & 0.519$\pm$0.07 & 80.8$\pm$0.4 & \textbf{85.4$\pm$0.2}/87.3$\pm$0.2 & 85.5$\pm$0.3/87.1$\pm$0.3 \\
 & Dual+Video & Merge & 0.507$\pm$0.09 & 80.7$\pm$0.2 & 83.8$\pm$0.3/87.3$\pm$0.1 & 84.1$\pm$0.5/87.2$\pm$0.4 \\
\bottomrule
\end{tabular}}
\label{tab:mosei-senti-apd}
\end{table*}

\begin{table*}[htbp]
\centering
\caption{Results on CMU MOSEI Emotion Recognition.}
\begin{tabular}{l|c|c|cccc}
\toprule
\multicolumn{3}{l|}{Model} & Acc. & F1 & Precision & Recall \\ \midrule
\multicolumn{3}{l|}{DFG \citep{mosei}} & 38.6 & 49.4 & 53.4 & 45.6  \\
\multicolumn{3}{l|}{MISA \citep{misa}} & 39.8 & 45.0 & 37.1 & 57.1 \\
\multicolumn{3}{l|}{RAVEN \citep{raven}} & 40.3 & 51.1 & \textbf{63.3} & 42.9 \\
\multicolumn{3}{l|}{HHMPN \citep{hhmpn}} & 43.4 & 52.8 & 59.1 & 47.6 \\
\multicolumn{3}{l|}{TAILOR \citep{tailor}} & 46.0 & 52.9 & 63.9 & 45.2 \\
\multicolumn{3}{l|}{SIMM \citep{simm}} & 41.8 & 48.4 & 48.2 & 48.6 \\
\multicolumn{3}{l|}{ML-GCN \citep{mlgcn}} & 43.7 & 52.4 & 57.3 & 48.2 \\ \midrule
\multirow{5}{*}{\textbf{i-Code}} & Pretrain Data & Fusion Attention & & & & \\ \cmidrule{2-7}
 & No Pretrain & Merge & 49.2$\pm$0.3	& 54.6$\pm$0.4 & 50.3$\pm$0.3 & 59.8$\pm$0.3 \\
 & Dual & Merge & 49.4$\pm$0.2  & 55.4$\pm$0.2 & 49.4$\pm$0.3 & \textbf{63.0$\pm$0.1} \\
 & Dual & Co & \textbf{50.2$\pm$0.1} & \textbf{56.2$\pm$0.2} & 50.7$\pm$0.2 & \textbf{63.0$\pm$0.2} \\
 & Video & Merge & 49.4$\pm$0.3 & 55.3$\pm$0.3 & 49.6$\pm$0.2 & 62.4$\pm$0.4 \\
 & Dual+Video & Merge & 49.8$\pm$0.2 & 56.0$\pm$0.1 & 50.8$\pm$0.2 & 62.1$\pm$0.3 \\
\bottomrule
\end{tabular}
\label{tab:mosei-emo-apd}
\end{table*}

\subsection{Comparison with Models Using Alignment Information on CMU-MOSEI}
\label{sec:alignment}
We list performance of models leveraging alignment information in \Cref{tab:mosei-emo-apd-align}. i-Code, w/o using the alignment information between transcripts and audio, can outperform many baseline models using that information.

\begin{table*}[htbp]
\centering
\caption{Comparison between models w/ and w/o using the alignment information of CMU MOSEI Emotion Recognition.}
\begin{tabular}{l|c|c|cccc}
\toprule
\multicolumn{3}{l|}{Model} & Acc. & F1 & Precision & Recall \\ \midrule
\multicolumn{7}{c}{Model w/o Alignment Information} \\ \midrule
\multicolumn{3}{l|}{DFG \citep{mosei}} & 38.6 & 49.4 & 53.4 & 45.6  \\
\multicolumn{3}{l|}{MISA \citep{misa}} & 39.8 & 45.0 & 37.1 & 57.1 \\
\multicolumn{3}{l|}{RAVEN \citep{raven}} & 40.3 & 51.1 & 63.3 & 42.9 \\
\multicolumn{3}{l|}{MuIT \citep{mult}} & 42.3 & 52.3 & 63.6 & 44.5 \\
\multicolumn{3}{l|}{HHMPN \citep{hhmpn}} & 43.4 & 52.8 & 59.1 & 47.6 \\
\multicolumn{3}{l|}{TAILOR \citep{tailor}} & 46.0 & 52.9 & \textbf{63.9} & 45.2 \\
\multicolumn{3}{l|}{SIMM \citep{simm}} & 41.8 & 48.4 & 48.2 & 48.6 \\
\multicolumn{3}{l|}{ML-GCN \citep{mlgcn}} & 43.7 & 52.4 & 57.3 & 48.2 \\ \midrule
\multirow{5}{*}{\textbf{i-Code}} & Pretrain Data & Fusion Attention & & & & \\ \cmidrule{2-7}
 & No Pretrain & Merge & 49.2	& 54.6 & 50.3 & 59.8 \\
 & Dual & Merge & 49.4  & 55.4 & 49.4 & \textbf{63.0} \\
 & Dual & Co & \textbf{50.2} & \textbf{56.2} & 50.7 & \textbf{63.0} \\
 & Video & Merge & 49.4 & 55.3 & 49.6 & 62.4 \\
 & Dual+Video & Merge & 49.8 & 56.0 & 50.8 & 62.1 \\ \midrule
\multicolumn{7}{c}{Model with Alignment Information} \\ \midrule
\multicolumn{3}{l|}{DFG \citep{mosei}} & 39.6 & 51.7 & 59.5 & 45.7  \\
\multicolumn{3}{l|}{RAVEN \citep{raven}} & 41.6 & 51.7 & 58.8 & 46.1 \\
\multicolumn{3}{l|}{MuIT \citep{mult}} & 44.5 & 53.1 & 61.9 & 46.5 \\
\multicolumn{3}{l|}{MISA \citep{misa}} & 43.0 & 50.9 & 45.3 & 57.1 \\
\multicolumn{3}{l|}{HHMPN \citep{hhmpn}} & 45.9 & 55.9 & 60.2 & 49.6 \\
\multicolumn{3}{l|}{SIMM \citep{simm}} & 43.2 & 52.5 & 56.1 & 49.5 \\
\multicolumn{3}{l|}{ML-GCN \citep{mlgcn}} & 41.1 & 50.9 & 54.6 & 47.6 \\
\bottomrule
\end{tabular}
\label{tab:mosei-emo-apd-align}
\end{table*}

\subsection{Results of Using Other Single-modality Encoders}
\label{sec:speech-encoder}
In this section, we train two i-Code models on dual datasets with the merge-attention mechanism, one with WavLM-large as the speech encoder and the other with HuBERT-large. The i-Code model with WavLM shows slightly better performance on MOSEI Emotion Recognition (\Cref{tab:mosei-speech-encoder}), and higher accuracy on the VIOLIN dataset (\Cref{tab:violin-speech-encoder}).

\begin{table*}[htbp]
\centering
\caption{Comparison of using different speech encoders (WavLM v.s. HuBERT) in i-Code on CMU MOSEI Sentiment Analysis.}
\begin{tabular}{l|c|c|cccc}
\toprule
\multicolumn{3}{l|}{Model} & MAE $(\downarrow)$ & Corr. & Acc-2 & F1 \\ \midrule
\multicolumn{3}{l|}{i-Code w/ WavLM}  & \textbf{0.502} & 81.1 & 85.3/\textbf{87.5} & \textbf{85.6}/\textbf{87.4} \\
\multicolumn{3}{l|}{i-Code w/ HuBERT} & 0.504 & \textbf{81.2} & 85.3/87.2	& 85.4/87.1 \\
\bottomrule
\end{tabular}
\label{tab:mosei-speech-encoder}
\end{table*}

\begin{table*}[htbp]
\centering
\caption{Comparison of using different speech encoders (WavLM v.s. HuBERT) in i-Code on VIOLIN dataset.}
\begin{tabular}{l|ccc}
\toprule
Model & i-Code w/ WavLM & i-Code w/ HuBERT & \\ \midrule
Accuracy &   72.9 & 72.1  \\
\bottomrule
\end{tabular}
\label{tab:violin-speech-encoder}
\end{table*}

\end{document}

%% file: math_commands.tex
\usepackage{amsmath,amsfonts,bm}

\newcommand{\innerproduct}[2]{\langle #1, #2 \rangle}

\def\eqref#1{equation~\ref{#1}}

\def\1{\bm{1}}

\def\vp{{\bm{p}}}

\def\vr{{\bm{r}}}
\def\vs{{\bm{s}}}
\def\vt{{\bm{t}}}
\def\vu{{\bm{u}}}
\def\vv{{\bm{v}}}

\DeclareMathAlphabet{\mathsfit}{\encodingdefault}{\sfdefault}{m}{sl}
\SetMathAlphabet{\mathsfit}{bold}{\encodingdefault}{\sfdefault}{bx}{n}

\def\gB{{\mathcal{B}}}

\def\gL{{\mathcal{L}}}

\newcommand{\R}{\mathbb{R}}



%% file: main.bbl
\begin{thebibliography}{62}
\providecommand{\natexlab}[1]{#1}
\providecommand{\url}[1]{\texttt{#1}}
\expandafter\ifx\csname urlstyle\endcsname\relax
  \providecommand{\doi}[1]{doi: #1}\else
  \providecommand{\doi}{doi: \begingroup \urlstyle{rm}\Url}\fi

\bibitem[Akbari et~al.(2021)Akbari, Yuan, Qian, Chuang, Chang, Cui, and
  Gong]{vatt}
Hassan Akbari, Liangzhe Yuan, Rui Qian, Wei-Hong Chuang, Shih-Fu Chang, Yin
  Cui, and Boqing Gong.
\newblock Vatt: Transformers for multimodal self-supervised learning from raw
  video, audio and text.
\newblock \emph{Advances in Neural Information Processing Systems}, 34, 2021.

\bibitem[Alayrac et~al.(2020)Alayrac, Recasens, Schneider, Arandjelovi{\'c},
  Ramapuram, De~Fauw, Smaira, Dieleman, and Zisserman]{alayrac2020self}
Jean-Baptiste Alayrac, Adria Recasens, Rosalia Schneider, Relja
  Arandjelovi{\'c}, Jason Ramapuram, Jeffrey De~Fauw, Lucas Smaira, Sander
  Dieleman, and Andrew Zisserman.
\newblock Self-supervised multimodal versatile networks.
\newblock \emph{Advances in Neural Information Processing Systems},
  33:\penalty0 25--37, 2020.

\bibitem[Arnab et~al.(2021)Arnab, Dehghani, Heigold, Sun, Lu{\v{c}}i{\'c}, and
  Schmid]{vivit}
Anurag Arnab, Mostafa Dehghani, Georg Heigold, Chen Sun, Mario Lu{\v{c}}i{\'c},
  and Cordelia Schmid.
\newblock Vivit: A video vision transformer.
\newblock In \emph{Proceedings of the IEEE/CVF International Conference on
  Computer Vision}, pp.\  6836--6846, 2021.

\bibitem[Baevski et~al.(2020)Baevski, Zhou, Mohamed, and Auli]{wav2vec2}
Alexei Baevski, Yuhao Zhou, Abdelrahman Mohamed, and Michael Auli.
\newblock wav2vec 2.0: A framework for self-supervised learning of speech
  representations.
\newblock \emph{Advances in Neural Information Processing Systems},
  33:\penalty0 12449--12460, 2020.

\bibitem[Chen \& Kong(2021)Chen and Kong]{gve}
Junwen Chen and Yu~Kong.
\newblock Explainable video entailment with grounded visual evidence.
\newblock In \emph{Proceedings of the IEEE/CVF International Conference on
  Computer Vision}, 2021.

\bibitem[Chen et~al.(2021)Chen, Wang, Chen, Wu, Liu, Chen, Li, Kanda, Yoshioka,
  Xiao, et~al.]{wavlm}
Sanyuan Chen, Chengyi Wang, Zhengyang Chen, Yu~Wu, Shujie Liu, Zhuo Chen, Jinyu
  Li, Naoyuki Kanda, Takuya Yoshioka, Xiong Xiao, et~al.
\newblock Wavlm: Large-scale self-supervised pre-training for full stack speech
  processing.
\newblock \emph{arXiv preprint arXiv:2110.13900}, 2021.

\bibitem[Chen et~al.(2019)Chen, Wei, Wang, and Guo]{mlgcn}
Zhao-Min Chen, Xiu-Shen Wei, Peng Wang, and Yanwen Guo.
\newblock Multi-label image recognition with graph convolutional networks.
\newblock In \emph{Proceedings of the IEEE/CVF conference on computer vision
  and pattern recognition}, pp.\  5177--5186, 2019.

\bibitem[Chung et~al.(2020)Chung, Zhu, and Zeng]{splat}
Yu-An Chung, Chenguang Zhu, and Michael Zeng.
\newblock Splat: Speech-language joint pre-training for spoken language
  understanding.
\newblock \emph{arXiv preprint arXiv:2010.02295}, 2020.

\bibitem[Devlin et~al.(2019)Devlin, Chang, Lee, and Toutanova]{bert}
Jacob Devlin, Ming-Wei Chang, Kenton Lee, and Kristina Toutanova.
\newblock {BERT}: Pre-training of deep bidirectional transformers for language
  understanding.
\newblock In \emph{Proceedings of the 2019 Conference of the North {A}merican
  Chapter of the Association for Computational Linguistics: Human Language
  Technologies, Volume 1 (Long and Short Papers)}, pp.\  4171--4186,
  Minneapolis, Minnesota, June 2019. Association for Computational Linguistics.
\newblock \doi{10.18653/v1/N19-1423}.
\newblock URL \url{https://aclanthology.org/N19-1423}.

\bibitem[Dong et~al.(2021)Dong, Bao, Zhang, Chen, Zhang, Yuan, Chen, Wen, and
  Yu]{peco}
Xiaoyi Dong, Jianmin Bao, Ting Zhang, Dongdong Chen, Weiming Zhang, Lu~Yuan,
  Dong Chen, Fang Wen, and Nenghai Yu.
\newblock Peco: Perceptual codebook for bert pre-training of vision
  transformers.
\newblock \emph{arXiv preprint arXiv:2111.12710}, 2021.

\bibitem[Dosovitskiy et~al.(2020)Dosovitskiy, Beyer, Kolesnikov, Weissenborn,
  Zhai, Unterthiner, Dehghani, Minderer, Heigold, Gelly, et~al.]{vit}
Alexey Dosovitskiy, Lucas Beyer, Alexander Kolesnikov, Dirk Weissenborn,
  Xiaohua Zhai, Thomas Unterthiner, Mostafa Dehghani, Matthias Minderer, Georg
  Heigold, Sylvain Gelly, et~al.
\newblock An image is worth 16x16 words: Transformers for image recognition at
  scale.
\newblock In \emph{International Conference on Learning Representations}, 2020.

\bibitem[Dou et~al.(2021)Dou, Xu, Gan, Wang, Wang, Wang, Zhu, Liu, Zeng,
  et~al.]{meter}
Zi-Yi Dou, Yichong Xu, Zhe Gan, Jianfeng Wang, Shuohang Wang, Lijuan Wang,
  Chenguang Zhu, Zicheng Liu, Michael Zeng, et~al.
\newblock An empirical study of training end-to-end vision-and-language
  transformers.
\newblock \emph{arXiv preprint arXiv:2111.02387}, 2021.

\bibitem[Ekman et~al.(1980)Ekman, Freisen, and Ancoli]{ekman}
Paul Ekman, Wallace~V Freisen, and Sonia Ancoli.
\newblock Facial signs of emotional experience.
\newblock \emph{Journal of personality and social psychology}, 39\penalty0
  (6):\penalty0 1125, 1980.

\bibitem[Engin et~al.(2021)Engin, Schnitzler, Duong, and Avrithis]{diagsumqa}
Deniz Engin, Fran{\c{c}}ois Schnitzler, Ngoc~QK Duong, and Yannis Avrithis.
\newblock On the hidden treasure of dialog in video question answering.
\newblock In \emph{Proceedings of the IEEE/CVF International Conference on
  Computer Vision}, pp.\  2064--2073, 2021.

\bibitem[Gao et~al.(2021)Gao, Zhang, Han, and Callan]{gradcache}
Luyu Gao, Yunyi Zhang, Jiawei Han, and Jamie Callan.
\newblock Scaling deep contrastive learning batch size under memory limited
  setup.
\newblock In \emph{Proceedings of the 6th Workshop on Representation Learning
  for NLP (RepL4NLP-2021)}, pp.\  316--321, 2021.

\bibitem[Garcia \& Nakashima(2020)Garcia and Nakashima]{roll}
Noa Garcia and Yuta Nakashima.
\newblock Knowledge-based video question answering with unsupervised scene
  descriptions.
\newblock In \emph{Proceedings of the European Conference on Computer Vision},
  2020.

\bibitem[Garcia et~al.(2020)Garcia, Otani, Chu, and Nakashima]{knowit}
Noa Garcia, Mayu Otani, Chenhui Chu, and Yuta Nakashima.
\newblock Knowit vqa: Answering knowledge-based questions about videos.
\newblock In \emph{Proceedings of the AAAI Conference on Artificial
  Intelligence}, volume~34, pp.\  10826--10834, 2020.

\bibitem[Han et~al.(2021)Han, Chen, Gelbukh, Zadeh, Morency, and Poria]{bbfn}
Wei Han, Hui Chen, Alexander Gelbukh, Amir Zadeh, Louis-philippe Morency, and
  Soujanya Poria.
\newblock Bi-bimodal modality fusion for correlation-controlled multimodal
  sentiment analysis.
\newblock In \emph{Proceedings of the 2021 International Conference on
  Multimodal Interaction}, pp.\  6--15, 2021.

\bibitem[Hasan et~al.(2019)Hasan, Rahman, Zadeh, Zhong, Tanveer, Morency, and
  Hoque]{urfunny}
Md~Kamrul Hasan, Wasifur Rahman, AmirAli~Bagher Zadeh, Jianyuan Zhong,
  Md~Iftekhar Tanveer, Louis-Philippe Morency, and Mohammed~Ehsan Hoque.
\newblock Ur-funny: A multimodal language dataset for understanding humor.
\newblock In \emph{Proceedings of the 2019 Conference on Empirical Methods in
  Natural Language Processing and the 9th International Joint Conference on
  Natural Language Processing (EMNLP-IJCNLP)}, pp.\  2046--2056, 2019.

\bibitem[Hazarika et~al.(2020)Hazarika, Zimmermann, and Poria]{misa}
Devamanyu Hazarika, Roger Zimmermann, and Soujanya Poria.
\newblock Misa: Modality-invariant and-specific representations for multimodal
  sentiment analysis.
\newblock In \emph{Proceedings of the 28th ACM International Conference on
  Multimedia}, pp.\  1122--1131, 2020.

\bibitem[He et~al.(2020)He, Liu, Gao, and Chen]{deberta}
Pengcheng He, Xiaodong Liu, Jianfeng Gao, and Weizhu Chen.
\newblock Deberta: Decoding-enhanced bert with disentangled attention.
\newblock In \emph{International Conference on Learning Representations}, 2020.

\bibitem[Hsu et~al.(2021)Hsu, Bolte, Tsai, Lakhotia, Salakhutdinov, and
  Mohamed]{hubert}
Wei-Ning Hsu, Benjamin Bolte, Yao-Hung~Hubert Tsai, Kushal Lakhotia, Ruslan
  Salakhutdinov, and Abdelrahman Mohamed.
\newblock Hubert: Self-supervised speech representation learning by masked
  prediction of hidden units.
\newblock \emph{IEEE/ACM Transactions on Audio, Speech, and Language
  Processing}, 29:\penalty0 3451--3460, 2021.

\bibitem[Huang et~al.(2019)Huang, Hayashi, Wu, Kameoka, and
  Toda]{huang2019voice}
Wen-Chin Huang, Tomoki Hayashi, Yi-Chiao Wu, Hirokazu Kameoka, and Tomoki Toda.
\newblock Voice transformer network: Sequence-to-sequence voice conversion
  using transformer with text-to-speech pretraining.
\newblock \emph{arXiv preprint arXiv:1912.06813}, 2019.

\bibitem[Jia et~al.(2021)Jia, Yang, Xia, Chen, Parekh, Pham, Le, Sung, Li, and
  Duerig]{align}
Chao Jia, Yinfei Yang, Ye~Xia, Yi-Ting Chen, Zarana Parekh, Hieu Pham, Quoc Le,
  Yun-Hsuan Sung, Zhen Li, and Tom Duerig.
\newblock Scaling up visual and vision-language representation learning with
  noisy text supervision.
\newblock In \emph{International Conference on Machine Learning}, pp.\
  4904--4916. PMLR, 2021.

\bibitem[Lake et~al.(2017)Lake, Ullman, Tenenbaum, and
  Gershman]{lake2017building}
Brenden~M Lake, Tomer~D Ullman, Joshua~B Tenenbaum, and Samuel~J Gershman.
\newblock Building machines that learn and think like people.
\newblock \emph{Behavioral and brain sciences}, 40, 2017.

\bibitem[Li et~al.(2021{\natexlab{a}})Li, He, and Feng]{dukg}
Guohao Li, Feng He, and Zhifan Feng.
\newblock A clip-enhanced method for video-language understanding.
\newblock \emph{arXiv preprint arXiv:2110.07137}, 2021{\natexlab{a}}.

\bibitem[Li et~al.(2020)Li, Chen, Cheng, Gan, Yu, and Liu]{hero}
Linjie Li, Yen-Chun Chen, Yu~Cheng, Zhe Gan, Licheng Yu, and Jingjing Liu.
\newblock Hero: Hierarchical encoder for video+ language omni-representation
  pre-training.
\newblock In \emph{Proceedings of the 2020 Conference on Empirical Methods in
  Natural Language Processing (EMNLP)}, pp.\  2046--2065, 2020.

\bibitem[Li et~al.(2021{\natexlab{b}})Li, Lei, Gan, Yu, Chen, Pillai, Cheng,
  Zhou, Wang, Wang, et~al.]{value}
Linjie Li, Jie Lei, Zhe Gan, Licheng Yu, Yen-Chun Chen, Rohit Pillai, Yu~Cheng,
  Luowei Zhou, Xin~Eric Wang, William~Yang Wang, et~al.
\newblock Value: A multi-task benchmark for video-and-language understanding
  evaluation.
\newblock In \emph{Thirty-fifth Conference on Neural Information Processing
  Systems Datasets and Benchmarks Track (Round 1)}, 2021{\natexlab{b}}.

\bibitem[Li et~al.(2019)Li, Yatskar, Yin, Hsieh, and Chang]{visualbert}
Liunian~Harold Li, Mark Yatskar, Da~Yin, Cho-Jui Hsieh, and Kai-Wei Chang.
\newblock Visualbert: A simple and performant baseline for vision and language.
\newblock \emph{arXiv preprint arXiv:1908.03557}, 2019.

\bibitem[Liang et~al.(2021)Liang, Lyu, Fan, Wu, Cheng, Wu, Chen, Wu, Lee, Zhu,
  et~al.]{multibench}
Paul~Pu Liang, Yiwei Lyu, Xiang Fan, Zetian Wu, Yun Cheng, Jason Wu,
  Leslie~Yufan Chen, Peter Wu, Michelle~A Lee, Yuke Zhu, et~al.
\newblock Multibench: Multiscale benchmarks for multimodal representation
  learning.
\newblock In \emph{Thirty-fifth Conference on Neural Information Processing
  Systems Datasets and Benchmarks Track (Round 1)}, 2021.

\bibitem[Liu et~al.(2020)Liu, Chen, Cheng, Gan, Yu, Yang, and Liu]{violin}
Jingzhou Liu, Wenhu Chen, Yu~Cheng, Zhe Gan, Licheng Yu, Yiming Yang, and
  Jingjing Liu.
\newblock Violin: A large-scale dataset for video-and-language inference.
\newblock In \emph{Proceedings of the IEEE/CVF Conference on Computer Vision
  and Pattern Recognition}, pp.\  10900--10910, 2020.

\bibitem[Liu et~al.(2021{\natexlab{a}})Liu, Lin, Cao, Hu, Wei, Zhang, Lin, and
  Guo]{swin}
Ze~Liu, Yutong Lin, Yue Cao, Han Hu, Yixuan Wei, Zheng Zhang, Stephen Lin, and
  Baining Guo.
\newblock Swin transformer: Hierarchical vision transformer using shifted
  windows.
\newblock In \emph{Proceedings of the IEEE/CVF International Conference on
  Computer Vision}, pp.\  10012--10022, 2021{\natexlab{a}}.

\bibitem[Liu et~al.(2021{\natexlab{b}})Liu, Ning, Cao, Wei, Zhang, Lin, and
  Hu]{videoswin}
Ze~Liu, Jia Ning, Yue Cao, Yixuan Wei, Zheng Zhang, Stephen Lin, and Han Hu.
\newblock Video swin transformer.
\newblock \emph{arXiv preprint arXiv:2106.13230}, 2021{\natexlab{b}}.

\bibitem[Liu et~al.(2018)Liu, Shen, Lakshminarasimhan, Liang, Zadeh, and
  Morency]{lmf}
Zhun Liu, Ying Shen, Varun~Bharadhwaj Lakshminarasimhan, Paul~Pu Liang,
  AmirAli~Bagher Zadeh, and Louis-Philippe Morency.
\newblock Efficient low-rank multimodal fusion with modality-specific factors.
\newblock In \emph{Proceedings of the 56th Annual Meeting of the Association
  for Computational Linguistics (Volume 1: Long Papers)}, pp.\  2247--2256,
  2018.

\bibitem[Lu et~al.(2019)Lu, Batra, Parikh, and Lee]{vilbert}
Jiasen Lu, Dhruv Batra, Devi Parikh, and Stefan Lee.
\newblock Vilbert: Pretraining task-agnostic visiolinguistic representations
  for vision-and-language tasks.
\newblock \emph{Advances in neural information processing systems}, 32, 2019.

\bibitem[Luo et~al.(2021)Luo, Ji, Huang, Wang, Ji, and Li]{scalevlad}
Huaishao Luo, Lei Ji, Yanyong Huang, Bin Wang, Shenggong Ji, and Tianrui Li.
\newblock Scalevlad: Improving multimodal sentiment analysis via multi-scale
  fusion of locally descriptors.
\newblock \emph{arXiv preprint arXiv:2112.01368}, 2021.

\bibitem[Miech et~al.(2019)Miech, Zhukov, Alayrac, Tapaswi, Laptev, and
  Sivic]{howto100m}
Antoine Miech, Dimitri Zhukov, Jean-Baptiste Alayrac, Makarand Tapaswi, Ivan
  Laptev, and Josef Sivic.
\newblock Howto100m: Learning a text-video embedding by watching hundred
  million narrated video clips.
\newblock In \emph{Proceedings of the IEEE/CVF International Conference on
  Computer Vision}, pp.\  2630--2640, 2019.

\bibitem[Monfort et~al.(2021)Monfort, Jin, Liu, Harwath, Feris, Glass, and
  Oliva]{smit}
Mathew Monfort, SouYoung Jin, Alexander Liu, David Harwath, Rogerio Feris,
  James Glass, and Aude Oliva.
\newblock Spoken moments: Learning joint audio-visual representations from
  video descriptions.
\newblock In \emph{Proceedings of the IEEE/CVF Conference on Computer Vision
  and Pattern Recognition (CVPR)}, pp.\  14871--14881, June 2021.

\bibitem[Peerzade et~al.(2018)Peerzade, Deshmukh, and
  Waghmare]{peerzade2018review}
Gulnaz~Nasir Peerzade, RR~Deshmukh, and SD~Waghmare.
\newblock A review: Speech emotion recognition.
\newblock \emph{Int. J. Comput. Sci. Eng}, 6\penalty0 (3):\penalty0 400--402,
  2018.

\bibitem[Radford et~al.(2021)Radford, Kim, Hallacy, Ramesh, Goh, Agarwal,
  Sastry, Askell, Mishkin, Clark, et~al.]{clip}
Alec Radford, Jong~Wook Kim, Chris Hallacy, Aditya Ramesh, Gabriel Goh,
  Sandhini Agarwal, Girish Sastry, Amanda Askell, Pamela Mishkin, Jack Clark,
  et~al.
\newblock Learning transferable visual models from natural language
  supervision.
\newblock In \emph{International Conference on Machine Learning}, pp.\
  8748--8763. PMLR, 2021.

\bibitem[Ramesh et~al.(2021)Ramesh, Pavlov, Goh, Gray, Voss, Radford, Chen, and
  Sutskever]{dall-e}
Aditya Ramesh, Mikhail Pavlov, Gabriel Goh, Scott Gray, Chelsea Voss, Alec
  Radford, Mark Chen, and Ilya Sutskever.
\newblock Zero-shot text-to-image generation.
\newblock In \emph{International Conference on Machine Learning}, pp.\
  8821--8831. PMLR, 2021.

\bibitem[Schank \& Abelson(1975)Schank and Abelson]{schank1975scripts}
Roger~C Schank and Robert~P Abelson.
\newblock Scripts, plans, and knowledge.
\newblock In \emph{IJCAI}, volume~75, pp.\  151--157, 1975.

\bibitem[Schneider et~al.(2019)Schneider, Baevski, Collobert, and
  Auli]{wav2vec}
Steffen Schneider, Alexei Baevski, Ronan Collobert, and Michael Auli.
\newblock wav2vec: Unsupervised pre-training for speech recognition.
\newblock In \emph{INTERSPEECH}, 2019.

\bibitem[Shin et~al.(2021)Shin, Mun, On, Kang, Han, and Kim]{cstar}
Minchul Shin, Jonghwan Mun, Kyoung-Woon On, Woo-Young Kang, Gunsoo Han, and
  Eun-Sol Kim.
\newblock Winning the iccv'2021 value challenge: Task-aware ensemble and
  transfer learning with visual concepts.
\newblock \emph{arXiv preprint arXiv:2110.06476}, 2021.

\bibitem[Singh et~al.(2021)Singh, Hu, Goswami, Couairon, Galuba, Rohrbach, and
  Kiela]{flava}
Amanpreet Singh, Ronghang Hu, Vedanuj Goswami, Guillaume Couairon, Wojciech
  Galuba, Marcus Rohrbach, and Douwe Kiela.
\newblock Flava: A foundational language and vision alignment model.
\newblock \emph{arXiv preprint arXiv:2112.04482}, 2021.

\bibitem[Su et~al.(2019)Su, Zhu, Cao, Li, Lu, Wei, and Dai]{vl-bert}
Weijie Su, Xizhou Zhu, Yue Cao, Bin Li, Lewei Lu, Furu Wei, and Jifeng Dai.
\newblock Vl-bert: Pre-training of generic visual-linguistic representations.
\newblock In \emph{International Conference on Learning Representations}, 2019.

\bibitem[Sun et~al.(2020)Sun, Sarma, Sethares, and Liang]{iccn}
Zhongkai Sun, Prathusha Sarma, William Sethares, and Yingyu Liang.
\newblock Learning relationships between text, audio, and video via deep
  canonical correlation for multimodal language analysis.
\newblock In \emph{Proceedings of the AAAI Conference on Artificial
  Intelligence}, volume~34, pp.\  8992--8999, 2020.

\bibitem[Tang et~al.(2021)Tang, Lei, and Bansal]{decembert}
Zineng Tang, Jie Lei, and Mohit Bansal.
\newblock Decembert: Learning from noisy instructional videos via dense
  captions and entropy minimization.
\newblock In \emph{Proceedings of the 2021 Conference of the North American
  Chapter of the Association for Computational Linguistics: Human Language
  Technologies}, pp.\  2415--2426, 2021.

\bibitem[Tsai et~al.(2019)Tsai, Bai, Liang, Kolter, Morency, and
  Salakhutdinov]{mult}
Yao-Hung~Hubert Tsai, Shaojie Bai, Paul~Pu Liang, J~Zico Kolter, Louis-Philippe
  Morency, and Ruslan Salakhutdinov.
\newblock Multimodal transformer for unaligned multimodal language sequences.
\newblock In \emph{Proceedings of the 57th Annual Meeting of the Association
  for Computational Linguistics}, pp.\  6558--6569, 2019.

\bibitem[Vaswani et~al.(2017)Vaswani, Shazeer, Parmar, Uszkoreit, Jones, Gomez,
  Kaiser, and Polosukhin]{vaswani2017attention}
Ashish Vaswani, Noam Shazeer, Niki Parmar, Jakob Uszkoreit, Llion Jones,
  Aidan~N Gomez, {\L}ukasz Kaiser, and Illia Polosukhin.
\newblock Attention is all you need.
\newblock \emph{Advances in neural information processing systems}, 30, 2017.

\bibitem[Wang et~al.(2018)Wang, Singh, Michael, Hill, Levy, and Bowman]{glue}
Alex Wang, Amanpreet Singh, Julian Michael, Felix Hill, Omer Levy, and Samuel
  Bowman.
\newblock Glue: A multi-task benchmark and analysis platform for natural
  language understanding.
\newblock In \emph{Proceedings of the 2018 EMNLP Workshop BlackboxNLP:
  Analyzing and Interpreting Neural Networks for NLP}, pp.\  353--355, 2018.

\bibitem[Wang et~al.(2021)Wang, Chen, Wu, Chen, Dai, Liu, Jiang, Zhou, and
  Yuan]{bevt}
Rui Wang, Dongdong Chen, Zuxuan Wu, Yinpeng Chen, Xiyang Dai, Mengchen Liu,
  Yu-Gang Jiang, Luowei Zhou, and Lu~Yuan.
\newblock Bevt: Bert pretraining of video transformers.
\newblock \emph{arXiv preprint arXiv:2112.01529}, 2021.

\bibitem[Wang et~al.(2019)Wang, Shen, Liu, Liang, Zadeh, and Morency]{raven}
Yansen Wang, Ying Shen, Zhun Liu, Paul~Pu Liang, Amir Zadeh, and Louis-Philippe
  Morency.
\newblock Words can shift: Dynamically adjusting word representations using
  nonverbal behaviors.
\newblock In \emph{Proceedings of the AAAI Conference on Artificial
  Intelligence}, volume~33, pp.\  7216--7223, 2019.

\bibitem[Wu et~al.(2019)Wu, Chen, Hu, Wang, Chang, Wang, and Zhang]{simm}
Xuan Wu, Qing-Guo Chen, Yao Hu, Dengbao Wang, Xiaodong Chang, Xiaobo Wang, and
  Min-Ling Zhang.
\newblock Multi-view multi-label learning with view-specific information
  extraction.
\newblock In \emph{IJCAI}, pp.\  3884--3890, 2019.

\bibitem[Yu et~al.(2021)Yu, Xu, Yuan, and Wu]{selfmm}
Wenmeng Yu, Hua Xu, Ziqi Yuan, and Jiele Wu.
\newblock Learning modality-specific representations with self-supervised
  multi-task learning for multimodal sentiment analysis.
\newblock In \emph{Proceedings of the AAAI Conference on Artificial
  Intelligence}, volume~35, pp.\  10790--10797, 2021.

\bibitem[Yuan et~al.(2021)Yuan, Chen, Chen, Codella, Dai, Gao, Hu, Huang, Li,
  Li, et~al.]{florence}
Lu~Yuan, Dongdong Chen, Yi-Ling Chen, Noel Codella, Xiyang Dai, Jianfeng Gao,
  Houdong Hu, Xuedong Huang, Boxin Li, Chunyuan Li, et~al.
\newblock Florence: A new foundation model for computer vision.
\newblock \emph{arXiv preprint arXiv:2111.11432}, 2021.

\bibitem[Zadeh et~al.(2017)Zadeh, Chen, Poria, Cambria, and Morency]{tfn}
Amir Zadeh, Minghai Chen, Soujanya Poria, Erik Cambria, and Louis-Philippe
  Morency.
\newblock Tensor fusion network for multimodal sentiment analysis.
\newblock In \emph{Proceedings of the 2017 Conference on Empirical Methods in
  Natural Language Processing}, pp.\  1103--1114, 2017.

\bibitem[Zadeh et~al.(2018)Zadeh, Liang, Poria, Cambria, and Morency]{mosei}
AmirAli~Bagher Zadeh, Paul~Pu Liang, Soujanya Poria, Erik Cambria, and
  Louis-Philippe Morency.
\newblock Multimodal language analysis in the wild: Cmu-mosei dataset and
  interpretable dynamic fusion graph.
\newblock In \emph{Proceedings of the 56th Annual Meeting of the Association
  for Computational Linguistics (Volume 1: Long Papers)}, pp.\  2236--2246,
  2018.

\bibitem[Zellers et~al.(2021)Zellers, Lu, Hessel, Yu, Park, Cao, Farhadi, and
  Choi]{merlot}
Rowan Zellers, Ximing Lu, Jack Hessel, Youngjae Yu, Jae~Sung Park, Jize Cao,
  Ali Farhadi, and Yejin Choi.
\newblock Merlot: Multimodal neural script knowledge models.
\newblock \emph{Advances in Neural Information Processing Systems}, 34, 2021.

\bibitem[Zellers et~al.(2022)Zellers, Lu, Lu, Yu, Zhao, Salehi, Kusupati,
  Hessel, Farhadi, and Choi]{reserve}
Rowan Zellers, Jiasen Lu, Ximing Lu, Youngjae Yu, Yanpeng Zhao, Mohammadreza
  Salehi, Aditya Kusupati, Jack Hessel, Ali Farhadi, and Yejin Choi.
\newblock Merlot reserve: Neural script knowledge through vision and language
  and sound.
\newblock \emph{arXiv preprint arXiv:2201.02639}, 2022.

\bibitem[Zhang et~al.(2021)Zhang, Ju, Zhang, Li, Li, Zhu, and Zhou]{hhmpn}
Dong Zhang, Xincheng Ju, Wei Zhang, Junhui Li, Shoushan Li, Qiaoming Zhu, and
  Guodong Zhou.
\newblock Multi-modal multi-label emotion recognition with heterogeneous
  hierarchical message passing.
\newblock In \emph{Proceedings of the AAAI Conference on Artificial
  Intelligence}, volume~1, 2021.

\bibitem[Zhang et~al.(2022)Zhang, Chen, Shen, and Wang]{tailor}
Yi~Zhang, Mingyuan Chen, Jundong Shen, and Chongjun Wang.
\newblock Tailor versatile multi-modal learning for multi-label emotion
  recognition.
\newblock \emph{arXiv preprint arXiv:2201.05834}, 2022.

\end{thebibliography}
